\documentclass[a4paper]{article}
\usepackage{INTERSPEECH2021}
\usepackage{hyperref}
\usepackage{url}
\usepackage{booktabs}       
\usepackage{amsfonts}       
\usepackage{nicefrac}       
\usepackage{microtype}      
\usepackage{xcolor}
\usepackage{subcaption}
\usepackage{amsmath,bm,multicol,multirow}
\usepackage{cleveref}
\usepackage{adjustbox}
\usepackage{pgfplotstable}
\usepackage{latexsym}
\usepackage{setspace}
\usepackage{todonotes}
\usepackage{placeins}
\usepackage{afterpage}

\usepackage{microtype}

\newcommand{\xlsrp}{XLS-R}
\newcommand{\xlsrpb}[1]{\xlsrp{} {(#1B)}}

\newcommand{\xenglish}{X $\rightarrow$ English}

\newcommand{\wvbert}{w2v-bert-51 (0.6B)}
\newcommand{\mslam}{mSLAM}
\newcommand{\mctc}{mSLAM}
\newcommand{\xtremes}{\mbox{XTREME-S}}
\newcommand{\mctcb}[1]{\mctc{} {(#1B)}}

\title{XTREME-S: Evaluating Cross-lingual Speech Representations}

\name{
Alexis Conneau$^{\bigtriangleup}$\email{aconneau@google.com},
Ankur Bapna$^{\bigtriangleup}$\email{ankurbpn@google.com},
Yu Zhang$^{\bigtriangleup}$\email{ngyuzh@google.com},
Min Ma$^{\bigtriangleup}$,
Patrick von Platen$^{\clubsuit}$\email{patrick@huggingface.co},
Anton Lozhkov$^{\clubsuit}$,
Colin Cherry$^{\bigtriangleup}$,
Ye Jia$^{\bigtriangleup}$,
Clara Rivera$^{\bigtriangleup}$,
Mihir Kale$^{\bigtriangleup}$,
Daan Van Esch$^{\bigtriangleup}$,
Vera Axelrod$^{\bigtriangleup}$,
Simran~Khanuja$^{\bigtriangleup}$,
Jonathan H. Clark$^{\bigtriangleup}$,
Orhan Firat$^{\bigtriangleup}$,
Michael Auli$^{\Box}$,\\
Sebastian Ruder$^{\bigtriangleup}$\email{ruder@google.com},
Jason~Riesa$^{\bigtriangleup}$\email{riesa@google.com},
Melvin Johnson$^{\bigtriangleup}$\email{\{aconneau,ankurbpn,ngyuzh,ruder,riesa,melvinp\}@google.com; patrick@huggingface.co}
}
\address{$^{\bigtriangleup}$~Google Research \ \ \ \ $^{\clubsuit}$~Hugging Face \ \ \ \ $^{\Box}$~Meta AI}

\newcommand{\insertresults}{
    \begin{table*}[bht!]
        \begin{center}
            \resizebox{\linewidth}{!}{
            \begin{tabular}[b]{l|ccc|c|cc|c|c}
            \toprule
            \multirow{2}{*}{ Model} & \multicolumn{3}{c|}{Speech recognition} & \multicolumn{1}{c|}{Speech translation} & \multicolumn{2}{c|}{Speech classification}  & \multicolumn{1}{c|}{Speech retrieval} & \multirow{2}{*}{ Avg} \\
            & Fleurs & MLS & VoxPopuli & CoVoST-2 & Fleurs-LID & Minds-14 & Fleurs-R5 & \\
            \midrule
            Metrics & WER & WER & WER & BLEU & Acc. & F1 & P@1 & - \\
            \midrule
            \xlsrpb{0.3} & -  & 12.8 & 12.8 & 13.2 & - & - & - &  \\ 
            \wvbert & 14.1 & 9.9 & 9.3 & 20.4 & 71.4 & 82.7 & - & 59.1 \\ 
            \midrule
            \mctcb{0.6} & 14.6 & 10.1 & 9.2 & 20.6 & 73.3 & 86.9 & - & 59.7  \\ 
            \bottomrule
            \end{tabular}
            }
            
            \caption{Table of results for XTREME-S. \label{tab:results}}
        \end{center}
      \vspace{-0.6cm}
    \end{table*}
}

\newcommand{\insertdata}{
    \begin{table*}[bht!]
        \begin{center}
            \resizebox{\linewidth}{!}{
            \begin{tabular}[t]{l|l|cccccclll}
            \toprule
            {\bf Task} & {\bf Corpus} & {\bf $\vert$ Train $\vert$} & {\bf $\vert$ Dev $\vert$} & {\bf $\vert$ Test $\vert$} & {\bf $\vert$ Lang. $\vert$} &  {\bf Fine-tune} & {\bf $\vert$ Eval $\vert$} & {\bf Task} &  {\bf Metric} & {\bf Domain} \\
                \midrule
                \multirow{3}{*}{Speech recognition} & FLEURS & 999h & 122h & 293h & 102 & Multi & 1 & ASR & CER & Read-speech \\
                & MLS & 80h & 10h & 10h & 8 & Multi & 1 & ASR & WER & Read-speech \\
                & VoxPopuli & 1300h & 240h & 240h & 14 & Multi & 1 & ASR & WER & Euro Parl \\
                \midrule
                \multirow{1}{*}{Speech translation} & CoVoST-2 & 566h & 144h & 153h & 21 & Multi & 1 & AST & BLEU & Read-speech \\
                \midrule
                \multirow{2}{*}{Speech classification} & FLEURS & 999h & 122h & 293h & 102 & Multi & 1 & LangID & Acc. & Read-speech \\
                 & Minds-14 & 2h & 1h & 1h & 14 & Multi & 1 & Intent Cl. & Acc. & E-banking \\
                \midrule
                \multirow{1}{*}{Speech retrieval} & FLEURS & 49h & 6h & 14h & 5 & Either & 1/5 & Mining & P@K & Read-speech \\
                \bottomrule
            \end{tabular}
            }
            \captionof{table}{Characteristics of the datasets in XTREME-S. We report the number of hours for each train, dev and test set, and the number of languages. We specify the type of fine-tuning (monolingual or multilingual), which coincides with the number of fine-tuning runs. We also include the task, the metric and the speech domain.
            \label{tab:data}}
        \end{center}
    \end{table*}
}

\newcommand{\insertlanguages}{
\begin{table*}[ht!]
\begin{center}
\captionof{table}{\textbf{Characteristics of the 102 languages in XTREME-S}, with their ISO codes, language families, estimated number of speakers in millions (\#S) and number of hours of labeled data for each dataset: Fleurs (FLRS), Multilingual LibriSpeech (MLS), Vox Populi (VP), CoVoST-2 (CV-2), and Minds-14 (M-14).
Languages are grouped geographically in Western Europe (WE), Eastern Europe (EE), Central-Asia/Middle-East/North-Africa (CMN), Sub-Saharan Africa (SSA), South Asia (SA), South-East Asia (SEA) and CJK languages. \label{tab:langs}}
\begin{tabular}[t]{|r|l|l|l|l|l|r|c|c|c|c|c|}
\toprule
Idx & Language & ISO 639-3 & ISO 639-1 & Family & Group & \#S & FLRS & MLS & VP & CV-2 & M-14 \\ 
\midrule
1 & Afrikaans & afr & af & Indo-European & SSA & 17 & $\simeq$10 &  & & & \\ 
2 & Amharic & amh & am & Afro-Asiatic & SSA & 22 & $\simeq$10 & &  & & \\ 
3 & Arabic & ara & ar & Afro-Asiatic & CMN & 180 & $\simeq$10 & & &  2 & \\ 
4 & Armenian & hye & hy & Indo-European & EE & 6 & $\simeq$10 & &  & & \\ 
5 & Assamese & asm & as& Indo-European & SA & 13 & $\simeq$10 & & & & \\ 
6 & Asturian & ast & - & Indo-European & WE & 0.6 & $\simeq$10 & &  & & \\ 
7 & Azerbaijani & azj & az & Turkic & CMN & 18 & $\simeq$10 & & &  & \\ 
8 & Belarusian & bel & be & Indo-European & EE & 3 & $\simeq$10 &  & & & \\ 
9 & Bengali & ben & bn & Indo-European & SA & 260 & $\simeq$10 & &  & & \\ 
10 & Bosnian & bos & bs& Indo-European & WE & 9 & $\simeq$10 & &  & & \\ 
11 & Bulgarian & bul & bg & Indo-European & EE & 7 &  $\simeq$10 &  & & & \\ 
12 & Burmese & mya & my & Sino-Tibetan & SEA & 33 & $\simeq$10 & & & & \\ 
13 & Cantonese Chinese & yue & - & Sino-Tibetan & CJK & 920 & $\simeq$10  & & & & \\ 
14 & Catalan & cat & ca & Indo-European & WE & 4 & $\simeq$10 & &  & 81 & \\ 
15 & Cebuano & ceb & - & Austronesian & SEA & 16 & $\simeq$10 & &  & & \\ 
16 & Croatian & hrv & hr & Indo-European & WE & 4 & $\simeq$10 & & 43 & & \\ 
17 & Czech & ces & cs & Indo-European & EE & 10 & $\simeq$10 & & 62 &  10 & 1 \\ 
18 & Danish & dan & da & Indo-European & WE & 5 & $\simeq$10 & & & &  \\ 
19 & Dutch & nld & nl & Indo-European & WE & 21 & $\simeq$10 & 10 & 53 & 2 & 2 \\ 
20 & English & eng & en & Indo-European & WE & 550 & $\simeq$10 & 10 & & 543 &  4 \\
21 & Estonian & est & et & Uralic & EE & 1 & $\simeq$10 & & 3 & 3 & \\ 
22 & Filipino (Tagalog) & tgl & tl & Austronesian & SEA & 22 & $\simeq$10  & & & & \\ 
23 & Finnish & fin & fi & Uralic & WE & 5 & $\simeq$10 & & 27 & & \\ 
24 & French & fra & fr & Indo-European & WE & 280 & $\simeq$10 & 10 & 211 & 180 & 1 \\ 
25 & Fula & ful & ff & Atlantic-Congo & SSA & 12 & $\simeq$10 & & & & \\ 
26 & Galician & glg & gl & Indo-European & WE & 2 & $\simeq$10 &  & & & \\ 
27 & Ganda & lug & lg & Atlantic-Congo & SSA & 4 & $\simeq$10 & &  & & \\ 
28 & Georgian & kat & ka & Kartvelian & EE & 4 & $\simeq$10 &  & & & \\ 
29 & German & deu & de & Indo-European & WE & 83 & $\simeq$10 & 10 & 282 & 119 & 2 \\ 
30 & Greek & ell & el &  Indo-European & WE & 13 & $\simeq$10 & & & &  \\ 
31 & Gujarati & guj & gu & Indo-European & SA & 56 & $\simeq$10 & &  & & \\ 
32 & Hausa & hau & ha &  Afro-Asiatic & SSA & 70 & $\simeq$10 & & &  & \\ 
33 & Hebrew & heb & he & Afro-Asiatic & CMN & 4 & $\simeq$10 & &  & & \\ 
34 & Hindi & hin & hi & Indo-European & SA & 320 & $\simeq$10 & &  & & \\ 
35 & Hungarian & hun & hu & Uralic & WE & 13 & $\simeq$10 & & 63 & & \\ 
36 & Icelandic & isl & is & Indo-European & WE &  0.3 & $\simeq$10 &  & & & \\ 
37 & Igbo & ibo & ig & Atlantic-Congo & SSA & 18 & $\simeq$10 & & &  & \\ 
38 & Indonesian & ind & id & Austronesian & SEA & 200 & $\simeq$10 &  & & 1 & \\ 
39 & Irish & gle & ga & Indo-European & WE & 0.2 & $\simeq$10 & &  & & \\ 
40 & Italian & ita & it & Indo-European & WE & 61 & $\simeq$10 & 10 & 91 & 28 & 3 \\ 
41 & Japanese & jpn & ja & Japonic & CJK & 130 & $\simeq$10 & & &  1 & \\ 
42 & Javanese & jav & jv & Austronesian & SEA & 85 & $\simeq$10 &  & & & \\ 
43 & Kabuverdianu & kea & - & Indo-European & WE & 0.9 & $\simeq$10 & & & & \\ 
44 & Kamba & kam & - & Atlantic-Congo & SSA & 4 & $\simeq$10 & &  & & \\ 
45 & Kannada & kan & kn & Dravidian & SA & 43 & $\simeq$10 & & &  & \\ 
46 & Kazakh & kaz & kk & Turkic & CMN & 11 & $\simeq$10 & & & &  \\ 
47 & Khmer & khm & km & Austro-Asiatic & SEA & 16 & $\simeq$10 &  & & & \\ 
48 & Korean & kor & ko & Koreanic & CJK & 52 & $\simeq$10 & & &  & 1 \\ 
49 & Kyrgyz & kir & ky & Turkic & CMN & 8 & $\simeq$10 & & & & \\ 
50 & Lao & lao & lo & Kra-Dai & SEA & 20 & $\simeq$10 & &  & & \\ 
51 & Latvian & lav & lv & Indo-European & EE & 2 & $\simeq$10 & &  & 2 & \\ 
\bottomrule
\end{tabular}
\end{center}
\end{table*}

\begin{table*}[ht!]
\begin{center}
\begin{tabular}[t]{|r|l|l|l|l|l|r|c|c|c|c|c|}
\toprule
Idx & Language & ISO 639-3 & ISO 639-1 & Family & Group & \#S & FLRS & MLS & VP & CV-2 & M-14 \\ 
\midrule
52 & Lingala & lin & ln & Atlantic-Congo & SSA & 15 & $\simeq$10 &  & & & \\ 
53 & Lithuanian & lit & lt & Indo-European & EE & 2 & $\simeq$10 & & 2 & & \\ 
54 & Luo & luo & - & Nilo-Saharan & SSA & 4 & $\simeq$10 & & & &  \\ 
55 & Luxembourgish & ltz & lb & Indo-European & WE & 0.4 & $\simeq$10 & & & & \\ 
56 & Macedonian & mkd & mk & Indo-European & EE & 1 & $\simeq$10  & & & & \\ 
57 & Malay & msa & ms & Austronesian & SEA & 80 & $\simeq$10 & &  & & \\ 
58 & Malayalam & mal & ml & Dravidian & SA & 77 & $\simeq$10 & & & & \\ 
59 & Maltese & mlt & mt & Afro-Asiatic & WE & 0.5 & $\simeq$10  & & & & \\
60 & Mandarin Chinese & cmn & - & Sino-Tibetan & CJK & 80 & $\simeq$10  & & & & 1 \\ 
61 & Maori & mri & mi & Austronesian & SEA & 0.2 & $\simeq$10 &  & & & \\ 
62 & Marathi & mar & mr & Indo-European & SA & 83 & $\simeq$10  & & & & \\ 
63 & Mongolian & mon & mn & Mongolic & CMN & 5 & $\simeq$10 &  & & 3 & \\ 
64 & Nepali & npi & ne & Indo-European & SA & 16 & $\simeq$10  & & & & \\ 
65 & Northern Sotho & nso & - & Atlantic-Congo & SSA & 14 & $\simeq$10 & & & & \\ 
66 & Norwegian & nob & nb & Indo-European & WE & 5 & $\simeq$10 & &  & & \\ 
67 & Nyanja & nya & ny & Atlantic-Congo & SSA & 12 & $\simeq$10 & & &  & \\ 
68 & Occitan & oci & oc & Indo-European & WE & 0.5 & $\simeq$10 & & &  & \\ 
69 & Oriya & ory & or & Indo-European & SA & 35 & $\simeq$10 & & & &  \\ 
70 & Oromo & orm & om & Afro-Asiatic & SSA & 24 & $\simeq$10 & & & &  \\ 
71 & Pashto & pus & ps & Indo-European & CMN & 13 & $\simeq$10 & & &  & \\ 
72 & Persian & fas & fa & Indo-European & CMN & 40 & $\simeq$10 & &  & 5 & \\ 
73 & Polish & pol & pl & Indo-European & EE & 38 & $\simeq$10 & 10 & 111 & & 3 \\ 
74 & Portuguese (Brazil) & por & pt & Indo-European & WE & 220 & $\simeq$10 & 10 & & 7 & 3 \\ 
75 & Punjabi & pan & pa & Indo-European & SA & 113 & $\simeq$10 & & & & \\ 
76 & Romanian & ron & ro & Indo-European & EE & 19 & $\simeq$10 & & 89 & & \\ 
77 & Russian & rus & ru & Indo-European & EE & 150 & $\simeq$10 & & & 16 & 1 \\ 
78 & Serbian & srp & sr & Indo-European & EE & 6 & $\simeq$10 & &  & & \\ 
79 & Shona & sna & sn & Atlantic-Congo & SSA & 9 & $\simeq$10 & &  & & \\ 
80 & Sindhi & snd & sd & Indo-European & SA & 68 & $\simeq$10 & &  & & \\ 
81 & Slovak & slk & sk & Indo-European & EE & 4 & $\simeq$10 & & 35 & & \\ 
82 & Slovenian & slv & sl & Indo-European & EE & 2 & $\simeq$10 & & 10 & 2 & \\ 
83 & Somali & som & so & Afro-Asiatic & SSA & 24 & $\simeq$10 & & & &  \\ 
84 & Sorani Kurdish & ckb & - & Indo-European & CMN &  7 & $\simeq$10 & & & & \\ 
85 & Spanish & spa & es & Indo-European & WE & 490 & $\simeq$10 & 10 & 166 & 97 & 2 \\ 
86 & Swahili & swh & sw & Atlantic-Congo & SSA & 24 & $\simeq$10 & & &  & \\ 
87 & Swedish & swe & sv & Indo-European & WE & 8 & $\simeq$10 & & & 2 & \\ 
88 & Tajik & tgk & tg & Indo-European & CMN & 8 & $\simeq$10 & & & & \\ 
89 & Tamil & tam & ta & Dravidian & SA & 76 & $\simeq$10 & & & 2 & \\ 
90 & Telugu & tel & te & Dravidian & SA & 82 & $\simeq$10 & & & & \\ 
91 & Thai & tha & th & Kra-Dai & SEA & 20 & $\simeq$10 & & & & \\ 
92 & Turkish & tur & tr & Turkic & CMN & 82 & $\simeq$10 & & & 2 & \\ 
93 & Ukrainian & ukr & uk & Indo-European & EE & 32 & $\simeq$10 & & & & \\ 
94 & Umbundu & umb & - & Atlantic-Congo & SSA & 6 & $\simeq$10 & & & & \\ 
95 & Urdu & urd & ur & Indo-European & SA & 120 & $\simeq$10 & & & & \\ 
96 & Uzbek & uzb & uz & Turkic & CMN & 57 & $\simeq$10 & & & & \\ 
97 & Vietnamese & vie & vi & Austro-Asiatic & SEA & 96 & $\simeq$10 & & & & \\ 
98 & Welsh & cym & cy & Indo-European & WE & 0.7 & $\simeq$10 & & & 1 & \\ 
99 & Wolof & wol & wo & Atlantic-Congo & SSA & 4 & $\simeq$10 & & & & \\ 
100 & Xhosa & xho & xh & Atlantic-Congo & SSA & 19 & $\simeq$10 & & & & \\ 
101 & Yoruba & yor & yo & Atlantic-Congo & SSA & 21 & $\simeq$10 & & & & \\ 
102 & Zulu & zul & zu & Atlantic-Congo & SSA & 11 & $\simeq$10 & & & & \\ 
\bottomrule
\end{tabular}
\end{center}
\end{table*}

}

\newcommand{\insertsc}{
\begin{table}[bht!]
        \begin{center}
            \caption{{\bf Speech Classification }- MINDS-14 speech intent classification and Fleurs speech language identification accuracy. \label{tab:sc}}
            \resizebox{0.9\linewidth}{!}{
            \begin{tabular}[hbt!]{l|c|c}
            \toprule
                {Model} & {Fleurs-LID} & {Minds-14} \\
                \midrule
                \midrule
                \multicolumn{3}{l}{\it Our work: Speech Only} \\
                \midrule
                \wvbert & 71.4 & 82.7 \\
                \midrule
                \multicolumn{3}{l}{\it Our work: Speech + Text} \\
                \midrule
                \mctcb{0.6}  & 73.3 & 86.9 \\
                \mctcb{2} & 77.7 & 86.6  \\
                \bottomrule
            \end{tabular}
            }
      \vspace{-0.4cm}
        \end{center}
    \end{table}
}

\newcommand{\insertvp}{
\begin{table*}[bht!]
\centering
\caption{\textbf{  Speech recognition } - VoxPopuli ASR results in terms of WER.}
\begin{tabular}{l|rrrrrrrr}
\toprule
& en & de & it & fr & es & pl & ro & hu  \\
\midrule
Labeled data & 543h & 282h & 91h & 211h & 166h & 111h & 89h & 63h \\
\midrule 
\multicolumn{8}{l}{\it Prior work \cite{babu2021xls}} \\
\midrule
\xlsrpb{0.3} & 10.2 & 13.0 & 19.2 & 12.6 & 9.8 & 9.6 & 7.9 & 11.6 \\
\xlsrpb{1}  &    8.8 &   11.5 &   15.1 &   10.8 &   8.2 &   7.7 &   7.3 &   9.6 \\
\midrule
\multicolumn{8}{l}{\it Our work: Speech-only} \\
\midrule
\wvbert & 7.2 & 9.0 & 15.8 & 9.2 & 8.6 & 6.5 & 7.6 & 8.4 \\
\midrule
\multicolumn{8}{l}{\it Our work: Speech + Text} \\
\midrule
\mctcb{0.6} & 7.1 & 8.9 & 15.6 & 9.3 & 8.6 & 6.5 & 8.5 & 8.1  \\
\mctcb{2} & 7.0	  & 8.7 & 15.4 & 9.4 & 8.4 & 6.4 & 7.8 & 8.4 \\
\bottomrule
\toprule
& nl & cs & sl & fi & hr & sk & Avg \\
\midrule
Labeled data & 53h & 62h & 10h & 27h & 43h & 35h & \\
\midrule 
\multicolumn{8}{l}{\it Prior work \cite{babu2021xls}} \\
\midrule
\xlsrpb{0.3} & 14.8 & 10.5 & 24.5 & 14.2 & 12.3 & 8.9 & 12.8\\
\xlsrpb{1}  &   12.5 &   8.7 &   19.5 &   11.3 &   10.0 &   7.1 &   10.6 \\
\midrule
\multicolumn{8}{l}{\it Our work: Speech-only} \\
\midrule
\wvbert & 10.5 & 7.0	& 15.8 & 9.3 & 9.1	& 6.0	& 9.3\\
\midrule
\multicolumn{8}{l}{\it Our work: Speech + Text} \\
\midrule
\mctcb{0.6} & 10.3 & 7.0 & 14.2 & 9.2 & 9.1 & 5.9 & 9.2\\
\mctcb{2} &10.5	& 6.8 & 15.1 & 8.7 & 9.1 & 6.0 & \textbf{9.1}\\
\bottomrule
\end{tabular}    
\label{tab:vp_asr}
\end{table*}
}

\newcommand{\insertfleursasr}{
 \begin{table}[bht!]
        \begin{center}
            \captionof{table}{\textbf{  Speech recognition }- Fleurs massively multilingual ASR baselines, reporting CER, by geographical group. Observe the discrepancy between European and African languages.
            \label{tab:fleurs}}
            \resizebox{\linewidth}{!}{
            \begin{tabular}[t]{l|ccccccc|c}
            \toprule
            {  Model} & WE & EE & CMN & SSA & SA & SEA & CJK & All \\
                \midrule
                \midrule
                \multicolumn{1}{l|}{Number of languages} & 25 & 16 & 12 & 20 & 14 & 11 & 4 & 102 \\
                \midrule
                \midrule
                \multicolumn{9}{l}{\it Our work: Speech-only, no LM} \\
                \midrule
                \wvbert  & 10.7 & 9.9 & 14.5 & 15.6 & 17.4 & 14.7 & 24.6 & 14.1 \\
                \midrule
                \multicolumn{9}{l}{\it Our work: Speech + Text, no LM} \\
                \midrule
                \mctcb{0.6} & 10.6 & 10.0 & 14.8 & 16.4 & 19.2 & 14.9 & 25.0 & 14.6 \\
                \bottomrule
            \end{tabular}
            }
        \end{center}
    \end{table}
}

\newcommand{\insertmls}{
\begin{table*}[bht!]
        \begin{center}
            \caption{\textbf{  Speech recognition }- Multilingual LibriSpeech (MLS) ASR baselines in 8 languages, reporting WER. \label{tab:mls}}
            \resizebox{0.8\linewidth}{!}{
            \begin{tabular}[hbt!]{l|cccccccc|c}
            \toprule
                {  Model} & {  en }& {  de }& {  nl }& {  fr }& {  es }& {  it }& {  pt} & {  pl} & {  Avg} \\
                \midrule
                \midrule
                \multicolumn{1}{l|}{Number of training hours} & 10 & 10 & 10 & 10 & 10 & 10 & 10 & 10 & -  \\
                \midrule
                \multicolumn{9}{l}{\it Prior work (monolingual fine-tuning)~\cite{babu2021xls}} \\
                \midrule
                XLS-R(0.3B) & 15.9 & 9.0 & 13.5 & 12.4 & 8.1 & 13.1 & 17.0 & 13.9 & 12.8 \\
                XLS-R(2B) & 14.0 & 7.6 & \textbf{11.8} & 10.0 & 6.9 & 12.1 & 15.6 & 9.8 & 11.0 \\
                \midrule
                \multicolumn{9}{l}{\it Our work: Speech Only (multilingual fine-tuning)} \\
                \midrule
                \wvbert & 12.7 & 7.0 & 12.6 & 8.9 & 5.9 & 10.3 & 14.6 & \textbf{6.9} & 9.9 \\
                \midrule
                \multicolumn{9}{l}{\it Our work: Speech + Text (multilingual fine-tuning)} \\
                \midrule
                \mctcb{0.6} & 13.3 & 7.0 & 12.5 & 9.7 & \textbf{5.5} & 10.5 & \textbf{14.1} & 8.5 & 10.1 \\
                \mctcb{2} & \textbf{11.9} & \textbf{6.6} & 12.4 & \textbf{8.5} & 5.8 & \textbf{9.8} & 15.2 & 7.7 & \textbf{9.7} \\
                \bottomrule
            \end{tabular}
            }
       \vspace{-0.4cm}
        \end{center}
    \end{table*}
}

\newcommand{\insertcovostshort}{
\begin{table}[hbt!]
\centering
\caption{\textbf{  Speech translation }- CoVoST 2 X$\rightarrow$En summarized results in BLEU. Full per-language results are available in the Appendix Table~\ref{tab:covost_xen_full}.}
\label{tab:covost_xen_short}
\resizebox{0.9\linewidth}{!}{ 
\begin{tabular}{l|rrrr}
\toprule
\xenglish{} & high & mid & low & all \\
\midrule 
\multicolumn{5}{l}{\it Prior work, mBART decoder init.~\cite{babu2021xls}} \\
\midrule
\xlsrpb{0.3}  & 30.6 & 18.9 & 5.1 & 13.2 \\
\xlsrpb{2}  & 36.1 &   27.7 &   15.1 &   22.1 \\
\midrule
\multicolumn{5}{l}{\it Our Work: Speech Only} \\
\midrule
\wvbert & 35.6 & 25.3 & 13.4 & 20.4 \\
\midrule
\multicolumn{5}{l}{\it Our Work: Speech + Text} \\
\midrule
\mctcb{0.6} & 35.5 & 25.2 & 13.7 & 20.6 \\
\mctcb{2} & 36.3 & 27.5 & 15.6 & 22.4 \\
\bottomrule
\end{tabular} 

}
\end{table}
}

\newcommand{\insertretrieval}{
\begin{table}[hbt!]
\centering
\caption{\textbf{  Speech retrieval }- FLEURS speech-text retrieval accuracy for English, Amharic, Hindi, Japanese and Yoruba. Target transcriptions are retrieved from pools of 100k in-language sentences from Wikipedia or CommonCrawl.}
\label{tab:retrieval}
\resizebox{0.9\linewidth}{!}{ 
\begin{tabular}{l|ccccc}
\toprule
Model & en & am & hi & ja & yo \\
\midrule 
\multicolumn{5}{l}{\it Speech-text transcription retrieval} \\
\midrule
\mctcb{0.6} & - & - & - & - & - \\
\midrule
\multicolumn{5}{l}{\it Speech-text translation retrieval } \\
\midrule
\mctcb{0.6} & NA & - & - & - & - \\
\bottomrule
\end{tabular} 

}
\end{table}
}

\newcommand{\insertcovost}{
\begin{table*}[hbt!]
\centering
\caption{\textbf{  Speech translation }- CoVoST 2 X$\rightarrow$En full results in BLEU.}
\label{tab:covost_xen_full}
\begin{tabular}{l|rrrr|rrrrr|rrr}
\toprule
& \multicolumn{4}{c|}{High-resource} & \multicolumn{5}{c|}{Mid-resource} & \multicolumn{3}{c}{Low-resource}\\
\midrule
\xenglish{} & fr & de & es & ca & fa & it & ru & pt & zh & tr & ar & et \\
Train Hours & 264h & 184h & 113h & 136h & 49h & 44h & 18h & 10h & 10h & 4h & 2h & 3h \\
\midrule 
\multicolumn{8}{l}{\it Prior work, mBART Decoder init.~\cite{babu2021xls}} \\
\midrule
\xlsrpb{0.3} & 32.9 & 26.7 & 34.1 & 28.7 & 5.9 & 29.0 & 26.4 & 28.3 & 4.9 & 4.6 & 3.0 & 3.5 \\
\xlsrpb{2} &   37.6 &   33.6 &   39.2 &   33.8 &   12.9 &   34.9 & 39.5 &   41.8 &   9.4 &   16.7 &   17.1 & 11.1 \\
\midrule
\multicolumn{8}{l}{\it Our Work: Speech Only} \\
\midrule
\wvbert & 36.9 & 33.1  & 38.9 & 33.5 & 5.8 & 34.9 & 41.8 & 36.1 & 8.0 & 8.8 & 13.7 & 17.4  \\
\midrule
\multicolumn{8}{l}{\it Our Work: Speech + Text} \\
\midrule
\mctcb{0.6} & 36.7 & 32.7 & 39.1 & 33.4 & 6.2 & 35.0 & 41.7 & 34.2 & 8.7 & 11.7 & 13.3 & 17.2  \\
\mctcb{2}   & 37.6 & 33.8 & 39.5 & 34.4	& 8.8 & 36.1 & 43.6	& 42.0 & 7.1 & 19.7	& 15.8 & 18.6	\\
\bottomrule
\end{tabular}    
\begin{tabular}{l|rrrrrrrrr|rrrr}
\toprule
 & \multicolumn{9}{c|}{Low-resource} & \multicolumn{4}{c}{Average}\\
\midrule
\xenglish{} &  mn & nl & sv & lv & sl & ta & ja & id & cy & high & mid & low & all \\
Train Hours & 3h & 7h & 2h & 2h & 2h & 2h & 2h & 2h & 2h	& \\
\midrule 
\multicolumn{8}{l}{\it Prior work~\cite{babu2021xls}} \\
\midrule
\xlsrpb{0.3}  & 0.4 & 22.0 & 10.3 & 6.0 & 6.6 & 0.2 & 0.6 & 1.4 & 2.5 & 30.6 & 18.9 & 5.1 & 13.2 \\
\xlsrpb{2}  &   1.6 &   31.7 &   29.6 &   19.5 &   19.6 & 0.5 &   3.5 &   16.5 &   14.0 &   36.1 &   27.7 &   15.1 &   22.1 \\
\midrule
\multicolumn{8}{l}{\it Our Work: Speech Only} \\
\midrule
\wvbert & 0.3 & 33.8 & 33.9 & 16.0 & 25.5 & 0.3 & 0.9 & 3.5 & 6.2 & 35.6 & 25.3 & 13.4 & 20.4 \\
\midrule
\multicolumn{8}{l}{\it Our Work: Speech + Text} \\
\midrule
\mctcb{0.6} & 0.5 & 32.5 & 32.1 & 18.6 & 25.0 & 0.3 & 1.7 & 3.7 & 6.8 & 35.5 & 25.2 & 13.7 & 20.6 \\
\mctcb{2}   & 0.3 & 34.4 & 35.5	& 22.8 & 29.2 & 0.3	& 1.7 & 4.7	& 4.4 & 36.3 & 27.5	& 15.6 & 22.4  \\
\bottomrule
\end{tabular} 
\end{table*}
}

\newcommand{\insertasr}{
\begin{table}[hbt!]
\centering
\caption{ \textbf{Speech Recognition} - Average Character Error Rate (CER) for Fleurs and average word error rate for the VoxPopuli and MLS-10Hr datasets. Per-language results can be found in Appendix Tables~\ref{tab:fleurs}, ~\ref{tab:mls} and ~\ref{tab:vp_asr} respectively.}
\label{tab:asr}
\resizebox{0.9\linewidth}{!}{ 
\begin{tabular}{l|rrrr}
\toprule
Model & Fleurs & MLS & VoxPop \\
\midrule 
\multicolumn{3}{l}{\it Prior work~\cite{babu2021xls}} \\
\midrule
\xlsrpb{0.3} & -  & 12.8 & 12.8\\ 
\xlsrpb{2}  & -  & 11.0 & - \\ 
\midrule
\multicolumn{3}{l}{\it Our work: Speech-only} \\
\midrule
\wvbert & 14.1 & 9.9 & 9.3 \\ 
\midrule
\multicolumn{3}{l}{\it Our work: Speech + Text} \\
\midrule
\mctcb{0.6} & 14.6  & 10.1 & 9.2 \\ 
\mctcb{2} & - & 9.7 & 9.1 \\ 

\bottomrule
\end{tabular}    
}
\end{table}
}

\newcommand{\insertfleursfull}{
\begin{table*}[hbt!]
\centering
\caption{\textbf{  FLEURS full ASR results }. We report per-language results for all geographical language groups. FLEURS is a dataset that is complete at more than 97\%. Some slight improvements and changes may be done in a 2nd version of the dataset (e.g. missing recordings, or replaced low-quality recordings). Updates will be made on our platform. We expect average results not to change significantly.}
\label{tab:fleurs_full}

\begin{tabular}{l|rrrrrrrrrrrrrrr}
\toprule
& \multicolumn{15}{c}{Western European}\\
\midrule
Language & ast & bs &  ca &  hr &  da &  nl &  en &  fi &  fr &  gl &  de &  el &  hu &  is & ga \\
\midrule 
\wvbert & 8.7 & 5.8 & 4.3 & 9.3 & 11.3 & 6.0 & 17.2 & 3.0 & 9.6 & 8.6 & 8.0 & 11.7 & 24.9 & 11.9 & 39.5 \\
\mctcb{0.6} & 7.5 & 5.1 & 4.7 & 8.5 & 14.0 & 6.8 & 16.3 & 3.4 & 9.7 & 8.7 & 5.7 & 12.0 & 18.1 & 12.8 & 40.5 \\
\end{tabular}  

\begin{tabular}{l|rrrrrrrrrrrrrrr}
\toprule
& \multicolumn{10}{c|}{Western European (WE)} & \multicolumn{5}{c}{Eastern European} \\
\midrule
Language & it &  kea &  lb &  mt &  nb &  oc &  pt &  es &  sv & cy & am & be & bg & cs & et \\
\midrule 
\wvbert & 2.6 & 4.9 & 19.4 & 17.3 & 5.8 & 11.7 & 4.2 & 3.7 & 7.6 & 11.1 & 17.2 & 9.1 & 4.8 & 10.3 & 3.1 \\
\mctcb{0.6} & 2.3 & 5.1 & 21.0 & 17.3 & 6.1 & 12.7 & 4.4 & 3.3 & 7.8 & 12.0 & 17.8 & 7.5 & 5.2 & 9.2 & 3.5 \\
\end{tabular}  

\begin{tabular}{l|rrrrrrrrrrrrrrr}
\toprule
& \multicolumn{11}{c|}{Eastern European (EE)} & \multicolumn{4}{c}{Central-Asia and}\\
\midrule
Language & ka & lv & lt & mk & pl & ro & ru & sr & sk & sl & uk & ar & az & he & kk \\
\midrule 
\wvbert & 30.7 & 4.4 & 12.8 & 11.8 & 5.0 & 8.0 & 5.6 & 11.6 & 4.9 & 7.9 & 21.4 & 10.5 & 12.7 & 37.2 & 6.5 \\
\mctcb{0.6} & 31.0 & 4.5 & 11.6 & 9.8 & 6.3 & 8.4 & 6.6 & 12.2 & 4.8 & 10.3 & 21.4 & 11.0 & 15.9 & 42.5 & 5.7 \\
\end{tabular}  

\begin{tabular}{l|rrrrrrrrrrrrrrr}
\toprule
& \multicolumn{8}{c|}{Middle-East and North-Africa (CMN)} & \multicolumn{7}{c}{Sub-Saharan Africa}\\
\midrule
Language & ky & mn & ps & fa & ckb & tg & tr & uz & af & am & ff & lg & ha & ig & kam \\
\midrule 
\wvbert & 8.3 & 15.2 & 20.4 & 15.7 & 15.1 & 7.1 & 8.5 & 16.8 & 9.5 & 17.2 & 27.8 & 12.4 & 9.8 & 18.1 & 13.5 \\
\mctcb{0.6} & 8.0 & 16.1 & 21.1 & 10.0 & 15.0 & 7.6 & 9.7 & 15.5 & 11.9 & 17.8 & 27.5 & 12.9 & 10.5 & 18.7 & 14.0 \\
\end{tabular}  

\begin{tabular}{l|rrrrrrrrrrrrrrr}
\toprule
& \multicolumn{13}{c|}{Sub-Saharan Africa (SSA)} & \multicolumn{2}{c}{South-Asia} \\
\midrule
Language & ln & luo & nso & ny & om & sn & so & sw & umb & wo & xh & yo & zu & as & bn \\
\midrule 
\wvbert & 6.1 & 7.0 & 11.7 & 11.5 & 21.7 & 16.6 & 21.3 & 19.4 & 13.1 & 17.8 & 23.9 & 23.3 & 9.8 & 13.7 & 9.4 \\
\mctcb{0.6} & 6.8 & 7.4 & 11.9 & 12.4 & 22.6 & 17.6 & 23.4 & 20.2 & 14.0 & 18.8 & 25.1 & 23.2 & 10.8 & 14.0 & 9.7 \\
\end{tabular}  

\begin{tabular}{l|rrrrrrrrrrrrrrr}
\toprule
& \multicolumn{12}{c|}{South-Asia (SA)} & \multicolumn{3}{c}{South-East Asia}\\
\midrule
Language & gu & hi & kn & ml & mr & ne & or & pa & sd & ta & te & ur & my & ceb & tl\\
\midrule 
\wvbert & 9.3 & 12.4 & 7.0 & 8.6 & 14.8 & 13.0 & 19.2 & 13.6 & 16.0 & 11.8 & 12.0 & 82.9 & 18.2 & 5.9 & 7.1 \\
\mctcb{0.6} & 9.6 & 15.2 & 9.6 & 12.2 & 18.9 & 14.8 & 20.7 & 15.2 & 20.8 & 13.2 & 12.3 & 83.1 & 18.8 & 6.2 & 7.6 \\
\end{tabular}  

\begin{tabular}{l|rrrrrrrrrrrrrrr}
\toprule
& \multicolumn{8}{c|}{South-East Asia (SEA)} & \multicolumn{4}{c|}{CJK} & & & \\
\midrule
Language & id & jv & km & lo & ms & mi & th & vi & yue & cmn & ja & ko & All & & \\
\midrule 
\wvbert & 5.2 & 7.0 & 29.9 & 38.1 & 8.6 & 10.3 & 18.6 & 14.2 & 37.0 & 22.2 & 37.7 & 21.7 & 14.1 \\
\mctcb{0.6} & 5.6 & 7.1 & 30.2 & 37.5 & 7.2 & 11.2 & 20.1 & 14.3 & 39.8 & 23.1 & 39.2 & 22.4
 & 14.6 &  \\
\bottomrule
\end{tabular}

\end{table*}
}

\begin{document}

\maketitle
\begin{abstract}
We introduce \xtremes, a new benchmark to evaluate universal cross-lingual speech representations in many languages. XTREME-S covers four task families: speech recognition, classification, speech-to-text translation and retrieval. Covering 102 languages from 10+ language families, 3 different domains and 4 task families, XTREME-S aims to simplify multilingual speech representation evaluation, as well as catalyze research in ``universal'' speech representation learning. This paper describes the new benchmark and establishes the first speech-only and speech-text baselines using XLS-R and mSLAM on all downstream tasks. We motivate the design choices and detail how to use the benchmark. Datasets and fine-tuning scripts are made easily accessible through the HuggingFace platform.\footnote{\small\url{https://hf.co/datasets/google/xtreme_s}}
\end{abstract}

\section{Introduction}
In the past two decades, the exploding amount of content on the Internet has led to a pressing urgency to build systems that can understand text, speech, and videos in all of the world’s approximately 6,900 languages. Making speech technology available in all languages is especially important to give speakers of under-represented languages an equal voice on the Internet, and the possibility to make their content and culture known outside of their language cluster. Building speech systems for such a large number of languages is especially challenging but recent advances in self-supervised learning (SSL) present great opportunities to achieve this goal.


Speech pre-training techniques like wav2vec 2.0~\cite{baevski2020wav} have emerged as the predominant approach for automatic speech recognition (ASR) and direct speech-to-text translation (ST), and have made speech models much more data efficient: ASR models can be learnt with as little as a few hours of labeled data~\cite{xu2021self,baevski2021unsupervised}. 
Multilingual pre-training helps build better representations for languages that lack unannotated data, and thus enables the same data-efficient strategies for low-resource languages. Approaches like XLS-R~\cite{conneau-etal-2020-unsupervised,babu2021xls}, for example, have shown particularly strong results on several tasks, including ASR on BABEL and multilingual LibriSpeech, and AST on CoVoST-2. Following a recent trend in natural language processing, the speech community has made these multilingual pre-trained models publicly available to accelerate research in multilingual speech understanding.

To support this rapid development and to make better speech technology available in all languages of the world, the community requires high-quality datasets and a unified evaluation benchmark that is shared across researchers and practitioners. There has been significant progress in the past few years towards building publicly available multilingual evaluation datasets for speech understanding~\cite{pratap2020mls,wang2021voxpopuli,wang2020covost2}. Many research studies have, however, designed models on different tasks, and evaluated on a small and often disparate set of languages. This makes comparisons across methods difficult, slows down the development of multilingual representations, and hinders the evaluation of the generalization capabilities of such pre-trained models. The goal of this paper is to structure the evaluation of multilingual speech representation learning.

To address these issues and incentivize the rapidly-evolving research on general-purpose multilingual speech representation learning, we introduce XTREME-S, the Cross-lingual Transfer Evaluation of Multilingual Encoders for Speech benchmark. XTREME-S builds on top of the XTREME series of evaluation benchmarks for text understanding, with XTREME~\cite{hu2020xtreme} and XTREME-R~\cite{ruder2021xtreme}, which specialize in the evaluation of multilingual text representations and have helped the community improve multilingual language understanding, with impressive performance improvements on a variety of tasks.\footnote{\url{https://sites.research.google/xtreme}}

XTREME-S is meant to be a more exhaustive, thorough and complete evaluation of learned speech representations. It covers 102 diverse languages spanning more than 10 language families and includes four different task families: recognition, translation, classification and retrieval. The seven downstream tasks of XTREME-S also cover various domains, from read-speech to parliamentary speech. It also includes a new general-purpose massively multilingual evaluation dataset dubbed Fleurs in all of the 102 languages.

\begin{figure*}[h!]
	\begin{center}
          \includegraphics[width=0.7\linewidth]{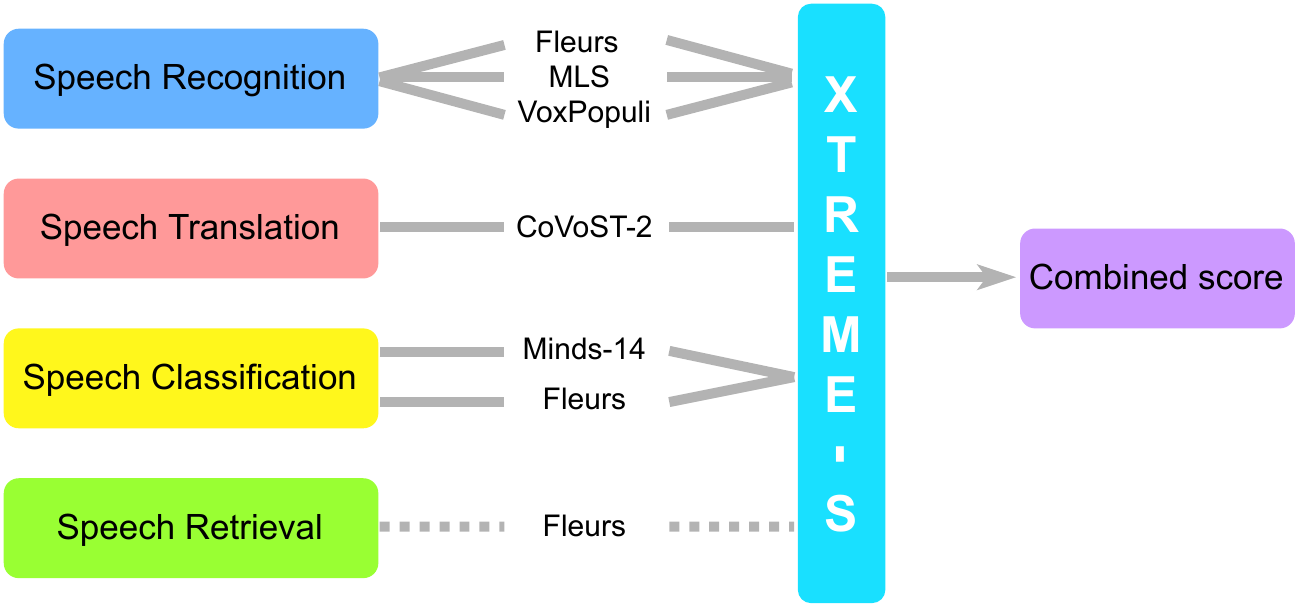}
        \captionof{figure}{\textbf{XTREME-S} is a benchmark for evaluating multilingual speech representation learning. It covers 4 task families, 3 speech domains and 102 diverse languages. Code and data publicly available at \url{https://hf.co/datasets/google/xtreme_s}.
        \label{fig:modelone}}
        \vspace{-0.2cm}
	\end{center}
\end{figure*}

\section{Related work}
\textbf{Multilingual representations}
Self-supervised learning methods like BERT~\cite{devlin-etal-2019-bert}, wav2vec 2.0~\cite{baevski2020wav} or w2v-BERT~\cite{chung2021w2v} have been extended to the cross-lingual setting through mBERT~\cite{devlin-etal-2019-bert}, XLM-R~\cite{conneau2020unsupervised} or XLS-R~\cite{conneau2021xlsr,babu2021xls}. These methods demonstrate the effectiveness of multilingual understanding in improving low-resource language representation through unsupervised cross-lingual transfer from higher-resource languages. 
Combined with the few-shot learning capability of wav2vec 2.0~\cite{xu2021self}, strong self-supervised speech representations can be built in low-resource languages, enabling training speech recognition systems with just a few hours of labeled data. XLS-R models demonstrate data-efficient capabilities in both speech recognition and speech translation for low-resource languages.
Recently, mSLAM ~\cite{bapna2022mslam} built a pre-trained multilingual model for both speech and text, leading to strong improvements on speech translation and even better data efficiency in low-resource languages. mSLAM is evaluated on text downstream tasks from XTREME~\cite{hu2020xtreme} and tasks from our new XTREME-S benchmark.
\\
\\
\textbf{Multilingual speech evaluation}
There has been a significant body of work on building trusted multilingual evaluation datasets for speech. IARPA introduced BABEL~\cite{gales2014babel} for evaluating speech models in low-resource languages. This dataset has been widely used in the speech community and covers real-world conversational telephone speech in 17 African and Asian low-resource languages. Recent work revived this dataset with different preprocessing \cite{alumae2016georgian,ragni2018confidence,inaguma2019transfer,conneau2021xlsr}. The CommonVoice effort~\cite{ardila2019common} offers a wide coverage of speech recognition data in more than 70 languages, with read speech of Wikipedia and other sentences. CommonVoice has been used namely for phoneme recognition~\cite{rivire2020unsupervised}. The Multilingual LibriSpeech~\cite{pratap2020mls} dataset extends the classical LibriSpeech task ~\cite{panayotov2015librispeech} to seven other European languages. VoxPopuli builds semi-supervised learning data from European Parliament session~\cite{wang2021voxpopuli} in 23 languages, and includes speech transcriptions and translations for 16 languages, as well as speech-to-speech translations. With more than 400k hours of unlabeled speech, VoxPopuli is also used as a public pre-training corpus~\cite{babu2021xls,bapna2022mslam}. In speech-to-text translation, CoVoST-2~\cite{wang2020covost2} has become one of the go-to datasets for multilingual evaluation, covering 21 language directions into English and English into 15 languages. Europarl-ST~\cite{jairsan2020a}, Must-C~\cite{di-gangi-etal-2019-must} and mTEDX~\cite{salesky2021multilingual} also provide common evaluation of speech translation. LangID can be evaluated using VoxLingua107~\cite{valk2020voxlingua107} on YouTube data in 107 languages, and CMU Wilderness~\cite{black2019cmu} on New Testament data in 700+ languages. Fleurs is a new multilingual speech understanding evaluation dataset in 102 languages.
%
\\
\\
\textbf{Multilingual benchmarks}
For text understanding, GLUE~\cite{wang2019glue} and SuperGLUE~\cite{wang2019superglue} provide common benchmarks for representation learning~\cite{devlin2018bert,yang2019xlnet,raffel2019exploring}. Methods like BERT, or T5 leverage GLUE to show the generalization ability of self-supervised learning on a variety of tasks. In the multilingual setting, new evaluation datasets like XNLI~\cite{conneau2018xnli}, MLQA~\cite{lewis2019mlqa} or TyDi~QA~\cite{tydiqa} are grouped in the XTREME benchmarks~\cite{hu2020xtreme,ruder2021xtreme}, on which methods like mBERT, XLM-R or mT5 show their generalization capabilities across languages. SUPERB ~\cite{yang2021superb} attempts to transpose GLUE to the speech setting, by grouping several common speech tasks to evaluate English speech models while LeBenchmark \cite{Evain2021} is designed for the evaluation of French self-supervised speech models. Our new XTREME-S benchmark groups several multilingual speech datasets and is the speech version of XTREME. The choice of tasks in XTREME-S is motivated by several factors explained in this work. Most tasks have been already used in previous work as evaluation for multilingual speech SSL.

\section{XTREME-S}
In this section, we describe the design decisions we made that led to the choice of tasks, domains and languages for our benchmark. Then we describe task families and their corresponding datasets.

\subsection{Design principles}
Given XTREME’s goal of providing an accessible benchmark for the evaluation of cross-lingual transfer learning on a
diverse and representative set of tasks and languages, we select the tasks and languages that make up the benchmark based on the following principles:
%
%
\\
\\
\textbf{Task difficulty}
Tasks should be sufficiently challenging that they are not saturated by the strongest existing baselines. The data should also be representative of the challenges faced by practitioners, under the constraint that the data should be publicly accessible.
\\
\\
\textbf{Diversity} We aim for task, domain and language diversity. Tasks should be diverse and cover several domains to provide a reliable evaluation of model generalization and robustness to noisy naturally-occurring speech in different environments. Languages should be diverse to ensure that models can adapt to a wide range of linguistic and phonological phenomena. Language coverage should not be unnecessarily large so as to avoid cumbersome evaluations. We note that the tasks are focused particularly on linguistic aspects of speech, while nonlinguistic/paralinguistic aspects of speech relevant to e.g. speech synthesis or voice conversion are not evaluated.
\\
\\
\textbf{Data efficiency}
The training sets of XTREME-S range from a few hours to a few hundred hours of labeled data per language. This is a few-shot setting suited for low-resource understanding. XTREME-S strongly encourages data-efficient self-supervised representation learning.
\\
\\
\textbf{Training efficiency}
Tasks should be trainable with a reasonable amount of time (few days) and compute (few GPUs). We enforce that constraint by having datasets focused on few-shot learning (e.g. Fleurs or MLS).
This is to make the benchmark accessible, in particular to practitioners working under resource constraints. We also minimize the number of required fine-tuning runs where we can, for instance by encouraging multilingual fine-tuning over monolingual fine-tuning.
\\
\\
\textbf{Monolingual data}
Unlabeled speech is available publicly through corpora already used in past work (e.g. MLS, VoxPopuli, CommonVoice). Unlabeled text data is available in all languages, for instance, through Common Crawl data as in the mC4 dataset\footnote{\url{https://www.tensorflow.org/datasets/catalog/c4\#c4multilingual}}. Speech data is however not abundant for all languages, so multilinguality is important to build strong representations for those languages. 
\\
\\
\textbf{Accessibility}
Each task should be available under a permissive license that allows the use and redistribution of the data for research purposes. When needed, we provide scripts to download and easily reproduce the preprocessing steps. Tasks have also been selected based on their usage by pre-existing multilingual pre-trained models, for simplicity.
\\
\\
\textbf{Reproducibility}
We encourage submissions that leverage publicly available speech and text datasets. Users should detail which data they use. In general, we encourage settings that can be reproduced by the community, but also encourage the exploration of new frontiers for speech representation learning.

\insertdata

\subsection{Tasks}
We present in this section the four task families of XTREME-S and their corresponding datasets.

\subsubsection{Speech Recognition (ASR)}
For speech recognition, we use three datasets: Fleurs, MLS and VoxPopuli, which cover more than 100 languages.
\\
\\
\textbf{Fleurs-ASR}
Fleurs
is the speech version of the FLoRes machine translation benchmark~\cite{goyal2021flores}. We use $2009$ n-way parallel sentences from the FLoRes dev and devtest publicly available sets, in $102$ languages. We collect between one and three recordings for each sentence ($2.3$ on average), and build new train-dev-test splits with $1509$, $150$ and $350$ sentences for train, dev and test respectively. Training sets have around 10 hours of supervision. Speakers of the train sets are different than speakers from the dev/test sets. Multilingual fine-tuning is used and "unit error rate" (characters, signs) of all languages is averaged. Languages and results are also grouped into seven geographical areas: Western Europe (WE), Eastern Europe (EE), Central-Asian/Middle-East/North-Africa (CMN), Sub-Saharan Africa (SSA), South Asia (SA), South-Eastern Asia (SEA) and CJK languages (CJK), as reported in Table~\ref{tab:langs}.
\\
\\
\textbf{MLS}
The Multilingual LibriSpeech (MLS) dataset is a large corpus derived from read audiobooks of Librivox and consists of 8 languages: \textit{Dutch (nl), English (en), French (fr), German (de), Italian (it), Polish (pl), Portuguese (pt), Spanish (es)}. The latest version of this corpus contains around 50k hours including 44k hours in English. The task consists of the official 10-hour splits provided by \cite{pratap2020mls}
to evaluate few-shot learning capabilities. We use multilingual fine-tuning on all languages at once.
\\
\\
\textbf{VoxPopuli}
VoxPopuli is a multilingual speech dataset for semi-supervised learning~\cite{wang2021voxpopuli}
. It contains 400k hours of unannotated speech as well as speech transcriptions and translations. We use the 14 languages with more than 10 hours of data from the ASR task. Models are fine-tuned on all 14 languages at once, ranging from 543 hours of supervision for English to 10 hours for Slovenian. Word Error Rate (WER) is reported. The language modeling data is provided by VoxPopuli.

\subsubsection{Speech Translation (ST)}
For speech translation, we use all the 21 language pairs into English from the CoVoST-2 dataset.
\\
\\
\textbf{CoVoST-2}
CoVoST-2
is a large-scale multilingual speech translation corpus covering translations from 21 languages into English. This represents the largest open dataset available to date from total volume and language coverage perspective. We consider all languages to English, grouped into high/mid/low labeled data directions. The task has been widely used in recent speech representation learning~\cite{babu2021xls,bapna2022mslam} and has been recently expanded to cover speech-to-speech translation \cite{jia2022cvss}.


\subsubsection{Speech classification}
For speech classification, we include LangID and intent classification. After hyperparameter tuning, we encourage reporting the average result over 5 random seeds.
\\
\\
\textbf{Fleurs-LangID}
We use Fleurs as a LangID dataset by using the same train, dev and test splits as used for ASR. We report over classification accuracy over the 102 languages.
\\
\\
\textbf{Minds-14}
MINDS-14~\cite{gerz2021multilingual} is an intent classification task from spoken data. It covers 14 intents extracted from the e-banking domain, with spoken examples in 14 language varieties. We merge monolingual datasets into a single multilingual dataset, with a $30$-$20$-$50$\% train-dev-test split. 

\subsubsection{Speech retrieval (Optional)}
For speech-text ASR retrieval, we use the Fleurs dataset in 5 languages. Because it is a new task, we mark it as optional.
\\
\\
\textbf{Fleurs} 
We define a new speech-text ASR retrieval task based on fixed-size embeddings. For each speech query embedding, the embedding of the correct text transcription should be retrieved using similarity search (e.g. cosine similarity), as in bitext mining~\cite{artetxe2019massively}. For each language, the pool of transcription candidates is augmented with 100k sentences from Wikipedia. We encourage the use of a ranking loss for fine-tuning. The average accuracy over the five languages should be reported. This is an optional new task.

\subsection{Languages}
Our 102 languages cover various language families and geographical locations (see Table~\ref{tab:langs}), from Western Europe/Americas, Eastern Europe, Central-Asia, Middle-East, North-Africa, Sub-Saharan Africa, South Asia, South-East Asia to CJK languages. We have 36 languages covered by at least two evaluation datasets. The language coverage provides a good estimate of the generalization ability of multilingual models.

\section{Results}
In this section, we describe our baselines and the corresponding results. We also comment on the specificities of each downstream task and offer remarks on how results can be improved.

\insertresults

\subsection{Baselines}
We present two baselines. The first is a 600M parameter speech-only pre-trained wav2vec-BERT model trained on 429k unlabeled data in 51 languages from VoxPopuli, MLS, CommonVoice and BABEL, similar to XLS-R. The second is the 600m parameter mSLAM speech-text pre-trained model that leverages the same speech data, as well more than 10TiB of unlabeled text data from mC4 and some ASR supervision. More details on these baselines, including fine-tuning details can be found in \cite{bapna2022mslam}. For some tasks, we also report results of the XLS-R models from \cite{babu2021xls}. If capacity constraints become an issue, we encourage practitioners to use same-capacity apples-to-apples comparisons with the smaller \xlsrpb{0.3} and \wvbert{} models.

\subsection{Speech recognition}
\insertasr
\insertfleursasr

In Table~\ref{tab:asr}, we report average character and word error rates on Fleurs, MLS and VoxPopuli. We see that \mslam{} obtains the best performance on MLS and VoxPopuli with 9.7 and 9.1 average WER.
%
 Pre-trained models obtain strong performance across domains and on both high-data regimes datasets like VoxPopuli as well as low-data regimes tasks like Fleurs and MLS. We observe in Table~\ref{tab:fleurs} that results are much better on the Western European group (with 11.5 average WER) than on other groups like Sub-Saharan African (26.7 average WER) or South Asian (20.7), which can be explained in part due to the larger amounts of unlabeled data in WE languages from MLS and VoxPopuli. Reducing the gaps across geographical groups is an important research direction for future work building on XTREME-S. Per-language results for MLS and VoxPopuli can be found in Appendix Tables ~\ref{tab:mls} and ~\ref{tab:vp_asr}.

\subsection{Speech translation}
\insertcovostshort

Average speech translation results are reported in Table~\ref{tab:covost_xen_short} and grouped by high-, mid- and low-resource languages. We observe that baselines perform well on different data regimes also significantly stronger on high-resource languages. Unlike previous approaches~\cite{wang2020covost2,li2021multilingual}, large-scale pre-trained multilingual models are able to obtain good performance on low-resource languages, showing again their few-shot capabilities in the case of speech translation. For most low-resource languages, only a couple of hours are available as supervision. Specifically, \wvbert{} obtains 13.4 and \mslam{} obtains 15.6 average BLEU on low-resource languages, 35.6 and 36.3 on high-resource languages. Overall, those models obtain 20.4 and 22.4 average BLEU respectively on all languages. On this dataset, only one multilingual fine-tuning run is done to simplify the evaluation. We encourage practitioners to also try different language re-sampling techniques, or various pre-training settings of the text decoder, as done for XLS-R. If using additional supervision, we still encourage reporting results which only leverage the supervision provided by the CoVoST-2 dataset.

\subsection{Speech classification}
\insertsc
We report our baselines on the two speech classification datasets in Table~\ref{tab:sc}. We see that the \mslam{} model obtains the best performance overall. Each of these datasets only require a single fine-tuning run; we build a multilingual training set from Minds-14 to reduce its inherent variance. Although not mandatory, we encourage the community to find the best hyperparameters for their fine-tuning setting, then re-run fine-tuning several times with different seeds, and report the average to minimize variance.

On Minds-14, mSLAM obtains around 86.6\% accuracy, and 77.7\% accuracy on Fleurs LangID, while \wvbert{} obtains 82.7 and 71.4 respectively. We note that on Fleurs-LangID, speakers are different between train sets and dev/test sets. Avoiding overfitting on speaker ID for the LangID task is essential for obtaining good performance. In general, speech classification tasks are prone to overfitting given the discrepancy between the richness of the input signal (speaker, domain, recording conditions) and the small number of output labels.

\insertretrieval  

\subsection{Speech retrieval (optional)}


Our speech-text ASR retrieval tasks consists of retrieving the correct transcription or English translation from an input speech utterance. We use the standard train/dev/test sets of the Fleurs data. The train set can be used for fine-tuning a siamese network with a pre-trained text and a pre-trained speech model. The [CLS] tokens of each model are used in the context of a ranking loss that is trained to match embeddings corresponding to speech-transcription pairs $(s,t)$ contrasted with negatives $s_c, t_c$: 
$$ max(0, \alpha \ -\ S(s,t) \ +\ 0.5*(S(s_c,t) + S(s,t_c))) $$
where $S(s,t)$ is a similarity measure of the speech and text embeddings $(s,t)$, e.g. the cosine similarity. At inference time, after models are fine-tuned with this ranking loss (or another), all embeddings of the dev and test sets are computed, as well as all the target text embeddings of 100k sentences from Wikipedia in corresponding language. The accuracy corresponds to the number of time the correct transcription/translation is  retrieved through nearest neighbor search from the pool of target sentences (which combine both the ground-truth dev/test transcriptions and the additional sentences from Wikipedia or CommonCrawl). Results on this task will be updated in the next version of the paper. The XTREME-S HuggingFace Dataset tool already provides the correct splits for this task. We hope this will create a new research path for speech search and speech retrieval.


\section{Discussion}
In this section, we discuss several components of the \mbox{XTREME-S} benchmark.
\\
\\
\textbf{On test sets: }
Test sets are available in open-source and are not hidden to the public. We trust practitioners to perform all hyperparameter search and checkpoint selection on the dev set, and eventually report performance on the test set. Results are however double-checked through the submission of the predictions of the model for each task.
\\
\\
\textbf{On speech data: }
We encourage the community to use similar unlabeled speech datasets across submissions when possible to encourage apple-to-apple comparisons across models. We do encourage submissions that also use different unlabeled speech, although preferably only in the case where there is a substantial difference (e.g. much smaller or much larger, or from more diverse sources, or using TTS-augmented data etc). Additional unlabeled speech data can be used for pre-training but also for self-training and other methods.
\\
\\
\textbf{On text data: }
The mC4 and Wikipedia datasets should cover all the languages of the XTREME-S benchmark, including low-resource ones. We encourage the use of these datasets for learning language models, for training text-augmented speech models, or using TTS augmentation for example. We hope the community can also develop smarter ways to adapt these very large unlabeled text datasets to each particular task and domain through filtering methods.
\\
\\
\textbf{On language modeling: }
The use of language model decoding is allowed. When using LMs, results should also be reported without LM fusion for comparison. The dataset and the type of LM used should be explicitly detailed in submissions and papers for reproducibility. When doing smart filtering of unlabeled text data, the technique should be explained clearly and the data released in open-source when possible.
\\
\\
\textbf{On the use of external supervision: }
At fine-tuning time, we ask that submissions leverage only the ASR supervision of each task. For instance, leveraging 10s of thousands of hours of ASR labeled data and then fine-tuning on MLS-10h English is not a valid submission. 
Submissions can potentially leverage all three datasets at once in a multi-task fashion (including during pre-training as in mSLAM). Additional unlabeled datasets can be used. For speech translation, additional supervision can be used in the form of open-sourced text-to-text machine translation data (e.g. from Opus) 
but any such data should be detailed explicitly in the submission and paper for clear comparisons to other methods. The TTS systems used to potentially augment the training set from the MT data should be reproducible.
For speech classification, the text data of each task can be used at training time but not at inference time. No other supervision is allowed. For speech retrieval, we encourage submissions to build generic universal fixed-size speech and text embeddings by leveraging all kinds of supervision (e.g. more ASR data). We only ask that new methods be easily reproduced (e.g. they do not use an unreasonable number of new datasets). In the exception of the exploration of very large-scale speech pre-training using proprietary data, which is encouraged and may be considered as a separate track, all extra supervision as well as unlabeled data should be easily accessible by other teams. The goal of the benchmark is not to prove that using more supervision leads to better performance but to discover new speech methods that lead to better data-efficient performance, in many languages. However, we believe giving more freedom in the submissions will lead to more interesting discoveries.
\\
\\
\textbf{On the average score: }
We weight differently each task of the XTREME-S benchmark. Speech recognition and translation each have a weight of 40\%, and speech classification has a weight of 20\%. The average score is computed in the following way: 
$$ 0.4 * \left(100 - \frac{\text{Fleurs} + \text{MLS} + \text{VP}}{3}\right)_{\text{(WER)}} + $$
$$ 0.4 * \text{CoVoST-2}_{\text{(BLEU)}} + 0.2 * \left(\frac{\text{F-LID} + \text{M-14}}{2}\right)_{\text{(Acc)}} $$
This is to give more importance to the core recognition and translation tasks.
\\
\\
\textbf{On submission: }
As previously mentioned, test sets are not hidden to the public. This means users can have access to their test results at the end of their hyperparameter tuning cycle on the dev sets. We ask users to be extra careful in this process not to inadvertently overfit on the test set. Additional test sets may be added in the future to confirm the generalization ability of submissions. We will provide a submission form where results can be double-checked for consistency before the submission is added to the leaderboard. More details will be added on the XTREME-S Dataset card\footnote{\small\url{https://hf.co/datasets/google/xtreme_s}}.

\section{Conclusion}
We presented XTREME-S, an evaluation benchmark meant to evaluate the generalization ability of multilingual speech pre-trained models. The benchmark consists of four key task types: recognition, translation, classification and retrieval. In total, XTREME-S covers 102 languages with various language families, from high-resource to low-resource, and different scripts. Tasks cover several domains and data regimes, from a few hours of supervision to more than a thousand hours, and are all directly open-sourced and made easily accessible. We presented two baselines: one speech-only pre-trained model and one speech-text pre-trained model that obtain strong results on each task. We believe there remains significant room for improvements on those tasks, in particular when it comes to reducing the gap between various language families or groups. We detailed in this paper the design choices of the XTREME-S benchmark and set guidelines for submissions. We also built a new dataset named Fleurs, in 102 languages, covering many low-resource languages. We hope XTREME-S will enable the community to build better speech representations in many languages, and enable rapid access to data-efficient speech technology for all the world's languages.

\bibliography{anthology,mybib}
\bibliographystyle{IEEEtran}

\cleardoublepage
\onecolumn  
\appendix
\insertlanguages

\insertcovost

\insertmls
\insertvp
\insertfleursfull

\end{document}